\documentclass[conference]{IEEEtran}
\IEEEoverridecommandlockouts

\usepackage{cite}
\usepackage{amsmath,amssymb,amsfonts}
\usepackage{algorithmic}
\usepackage{graphicx}
\usepackage{textcomp}
\usepackage{xcolor}
\usepackage{hyperref}
\usepackage{booktabs}
\usepackage{multirow}

\def\BibTeX{{\rm B\kern-.05em{\sc i\kern-.025em b}\kern-.08em
    T\kern-.1667em\lower.7ex\hbox{E}\kern-.125emX}}

\begin{document}

\title{A Standard Processing Pipeline for High-accuracy Measurement of Few-shot Regression on Laser Induced Breakdown Spectroscopy}
\author{\IEEEauthorblockN{1\textsuperscript{st} Hao Li}
	\IEEEauthorblockA{\textit{Electrical and Computer Engineering} \\
		\textit{University of Arizona}\\
		Tucson, USA \\
		\texttt{lihao@arizona.edu}}
}

\maketitle

\begin{abstract}
Laser-induced breakdown spectroscopy (LIBS) faces challenges in high-accuracy quantitative measurement under few-shot scenarios due to spectral noise and data scarcity. Traditional preprocessing methods often fail to preserve subtle spectral features or capture nonlinear correlations. This work proposes a standardized processing pipeline integrating diffusion-based denoising, attention-based autoencoder for dimensionality reduction, group shuffling data augmentation, and ordinary least squares regression. The diffusion module employs a 3D UNet architecture to remove spectral noise while preserving essential emission features. The attention-autoencoder captures nonlinear spectral correlations, effectively reducing high-dimensional spectral data to compact latent representations. Group shuffling data augmentation enhances model robustness by creating synthetic samples through feature group permutation. Experimental results on multiple elemental concentrations demonstrate that our Diffusion-DA-AE pipeline achieves superior performance with a mean RMAE of 0.2847, representing 37.7\% and 37.6\% improvements over baseline autoencoder and traditional PCA-PLS regression, respectively. The framework's effectiveness validates its generalizability and establishes a new benchmark for few-shot LIBS regression.
\end{abstract}


\section{Introduction}
\label{sec:intro}
Laser-induced breakdown spectroscopy (LIBS) has emerged as a powerful tool for rapid elemental and compositional analysis across diverse materials, leveraging its ability to generate characteristic emission spectra from plasma-induced ablation. However, high-accuracy quantitative measurement via LIBS is often hindered by two critical challenges: spectral noise (stemming from plasma instability, environmental interference, or detector limitations) and data scarcity (i.e., limited access to standard samples for model calibration). Traditional LIBS modeling approaches rely on large datasets to mitigate these issues, but in many practical scenarios—such as rare material analysis, on-site industrial monitoring, or hazardous substance detection—acquiring abundant standard samples is infeasible.

To address the gap in few-shot regression for LIBS, this work proposes a standardized processing pipeline tailored to high-accuracy measurement with limited training data (fewer than 20 samples). The pipeline integrates advanced denoising and dimensionality reduction techniques to enhance spectral quality and extract robust features, even when sample sizes are constrained. By focusing on generalizable data processing rather than material-specific adjustments, the framework aims to support quantitative analysis across a broad range of materials, extending LIBS’s applicability to resource-constrained environments.

Existing few-shot LIBS methods often prioritize either noise reduction or feature selection in isolation, leading to suboptimal performance when both challenges coexist. The traditional data preprocessing workflow, for instance, includes effective data selection based on mean and standard deviation thresholds, background removal via window-moving segmentation , normalization using spectral background areas, and feature extraction targeting atomic peaks and molecular bands, coupled with PCA for dimensionality reduction. While these steps lay a foundational framework, they have inherent limitations: the fixed threshold for data selection may inadvertently exclude spectra with valuable weak features; window-moving background removal can blunt sharp emission peaks; and PCA, as a linear method, fails to capture nonlinear correlations between spectral features (e.g., interdependencies between carbon and ash-related emission lines).

In contrast, our pipeline combines diffusion-based denoising and auto-encoder-driven dimensionality reduction to overcome these limitations. Diffusion denoising outperforms traditional noise-handling steps by iteratively refining noise while preserving subtle features: unlike the rigid thresholding in effective data selection, it adaptively distinguishes noise from weak emission lines (e.g., CN bands or minor element peaks), ensuring critical spectral information is retained. For dimensionality reduction, auto-encoders surpass PCA by learning nonlinear latent representations, capturing complex relationships between spectral features that linear methods miss—such as the inverse correlation between carbon and ash emission intensities. This synergy ensures that even with small datasets, the pipeline retains more predictive information than traditional preprocessing, enabling stable and accurate regression.

Our contribution summarizes as follows: 
\begin{itemize}
	\item This work is the first work trying to establish a standardized pipeline for few-shot regression that balances noise robustness and feature preservation, which contains diffusion for denoising, attention-autoencoder for dimension reduction, group shuffling for data augmentation, and OLS for regression.
	\item The proposed processing pipeline demonstrates superior performance across many elemental concentrations, achieving the lowest RMAE  compared to traditional methods and baseline approaches, validating its effectiveness for quantitative LIBS analysis.
\end{itemize}

\section{Preliminaries}
\subsection{Denoising}
Spectroscopic data are vulnerable to diverse noise sources, including detector readout noise, environmental fluctuations, cosmic ray artifacts, and radiation-based distortions like bremsstrahlung—all of which demand tailored denoising strategies\cite{yan2025review}. Bremsstrahlung—continuous radiation from decelerated charged particles—creates a broad, low-intensity background in high-energy spectra (e.g., XRF, LIBS), overlapping weak analyte signals. Denoising here requires suppressing such low-frequency noise while preserving sharp features. Traditional methods like polynomial fitting and AsLS effectively model and subtract bremsstrahlung-induced baselines, isolating target peaks\cite{eilers2003perfect}. For instance, AsLS-based correction in LIBS alloy analysis reduces bremsstrahlung interference, boosting trace element detectability\cite{jiang2021baseline}. High-frequency noise from detector electronics is often suppressed using Savitzky-Golay filtering (SGF), which employs local polynomial fitting to smooth data without blurring sharp peaks \cite{savitzky1964smoothing}. For impulsive noise like cosmic ray spikes—abrupt, narrowband artifacts—wavelet shrinkage denoising (WSD) decomposes spectra into multi-scale components, enabling targeted removal of spike-related coefficients while preserving genuine features \cite{donoho1994ideal}. In dynamic scenarios, such as real-time LIBS monitoring, Kalman filtering adaptively suppresses time-varying noise, including bremsstrahlung fluctuations, by recursively updating signal estimates\cite{kalman1960new}.

Traditional denoising methods, while effective for certain types of noise, face limitations when dealing with complex noise scenarios. However, these methods exhibit significant limitations. Li et al.\cite{shen2022single} pointed out that methods like wavelet transform (WT), Savitzky-Golay smoothing (SG), and asymmetric least squares (AsLS) can only reduce single interference items in Raman spectra, and a series of cumbersome trials are needed to complete preprocessing based on these traditional methods, with each solution only applicable to a specific dataset. Similarly, in the context of stellar spectra, the traditional wavelet-based denoising approaches require users to determine parameters such as the number of decomposition levels and noise estimates through trial-and-error, which is highly data-dependent\cite{gilda2019automatic}.

Advanced deep learning methods have revolutionized denoising in spectroscopy particularly when dealing with complex noise mixtures. Convolutional neural networks\cite{kazemzadeh2022cascaded,jiao2024three,liu2025learning,li2021latency,xu2017reliable,he2023new,shang2024study} can automatically learn hierarchical features from spectral data. Enhanced generative adversarial network\cite{he2023new} combined with transfer learning  achieves high accuracy in handling highly variable baselines in alloys and soils and shows good adaptability between different samples through transfer learning.

Notably, despite the significant progress in spectral denoising, current research efforts have yet to extensively explore the application of diffusion models for this purpose. Diffusion models, which have shown remarkable success in various data generation and denoising tasks, operate by learning the forward diffusion process of adding noise to data and then reversing it to recover the original clean data. In the context of image denoising, for instance, diffusion models have demonstrated the ability to handle complex noise distributions and generate high-quality denoised images\cite{he2023tdiffde}, as seen in studies such as the application of denoising diffusion space-spectral model (DDS2M) for hyperspectral image restoration\cite{miao2023dds2m}. This model effectively denoises hyperspectral images by inferring variational space - spectral module parameters during the reverse diffusion process, highlighting the potential of diffusion models in handling data with complex spatial and spectral characteristics.

\subsection{Dimension Reduction}
LIBS generates high-dimensional data with redundancy and noise, making dimensionality reduction critical—linear methods like PCA project data onto variance-capturing orthogonal components to simplify tasks, though they struggle with non-linear relationships, leading to adoption of non-linear techniques such as RBMs that model complex correlations via latent variables for enhanced low-dimensional representation\cite{vrabel2020restricted}. Feature selection and correction-based methods, such as Orthogonal Signal Correction (OSC) that removes components orthogonal to target variables, further aid dimensionality reduction while preserving key information. Feature selection strategies like spectral line screening and stepwise combinations of standard deviation filtering and Random Forest have optimized LIBS classification accuracy\cite{yuan2021rapid,ma2023step}.
Meanwhile, there exists a body of research demonstrating the efficacy of SFS and VAE, mainly confined to the domains of engineering and bioinformatics\cite{qiao2018protein,li2026golden,li2026revisiting,li2026r,ding2018hybrid,wei2020variations}. VAE was employed as a dimensionality reduction technique for Laser-Induced Breakdown Spectroscopy (LIBS) data of soil samples\cite{harefa2021performing}. 

\subsection{LIBS Regression}
Linear regression models like Multiple Linear Regression, which assume a linear relationship between spectral intensities and elemental concentrations, were early applied in LIBS for soil analysis but are limited by high collinearity in LIBS data. Partial Least Squares Regression (PLS) addresses this by extracting latent variables to handle high-dimensional and collinear data, with successful use in analyzing geological samples' LIBS spectra for elemental composition prediction\cite{li2023high,li2023real,li2026multi,zhang2024determination}. 
Non-linear regression methods like SVR, with kernel functions such as radial basis function, capture complex non-linear relationships, as seen in improving quantitative accuracy in determining fly ash carbon content via LIBS with matrix effect correction. ANNs (e.g., MLPs) and deep learning models excel in learning intricate LIBS spectral patterns\cite{van2023deep,ctvrtnickova2009}, outperforming traditional methods in tasks like predicting alloying element concentrations in post-consumer aluminum scrap sorting, while Ridge, LASSO, and Elastic net regression handle multicollinearity and variable selection in LIBS analysis of rock samples\cite{boucher2015study,shahani2022machine}.

\begin{figure*}[htbp]
	\centering
	\includegraphics[width=0.8\textwidth]{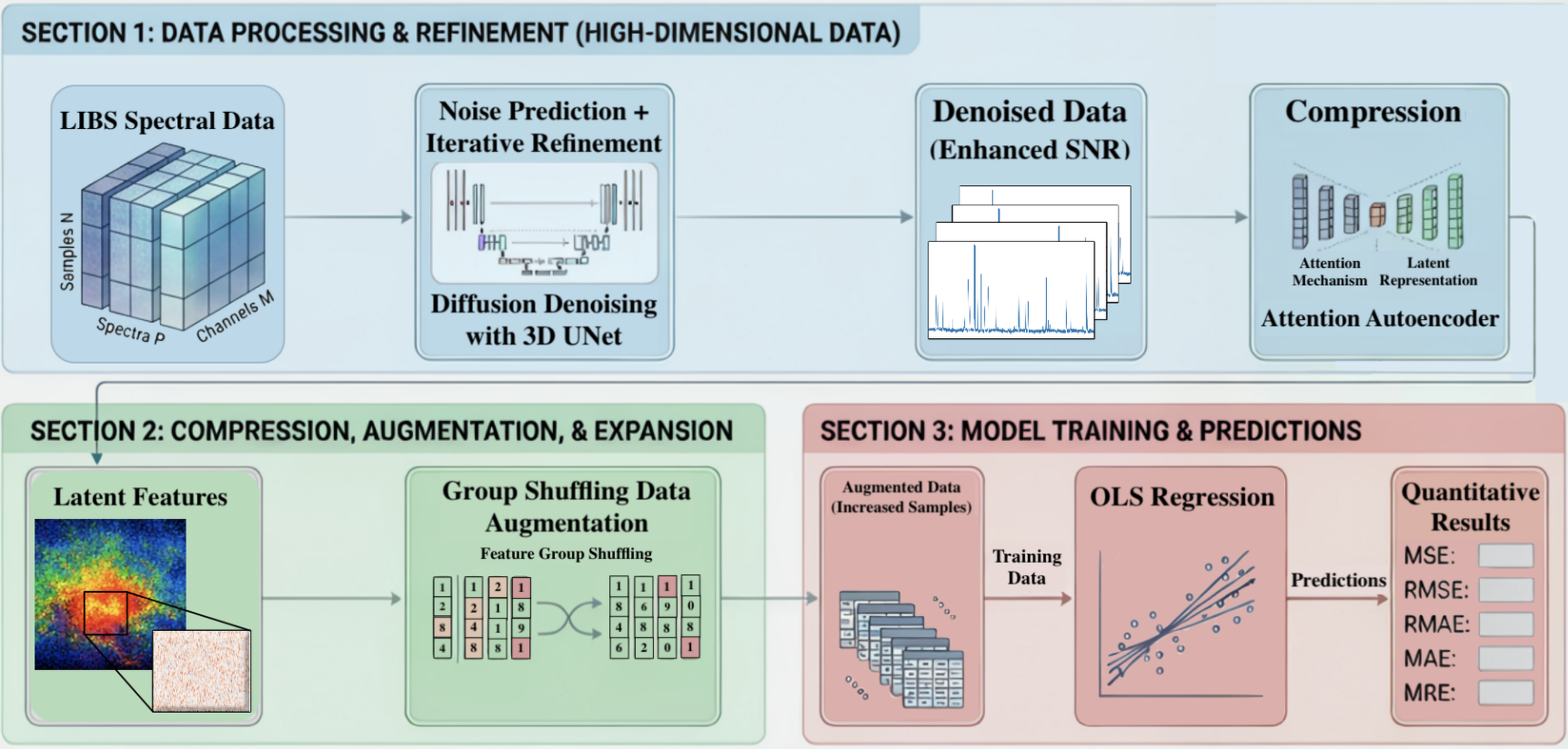}
	\caption{Overview of the proposed LIBS processing pipeline.}
	\label{fig:detailed_modules}
\end{figure*}
\section{Proposed Model}
\subsection{Model Overview}
\begin{figure*}[htbp]
	\centering
	\includegraphics[width=0.9\textwidth]{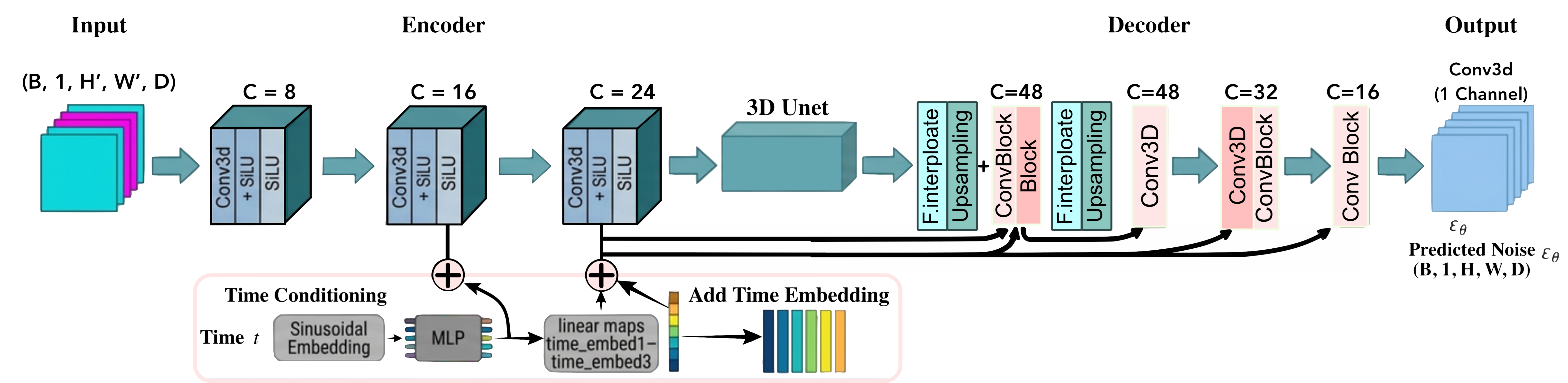}
	\caption{
		Schematic architecture of the 3D UNet used in the diffusion model. The network consists of an encoder, a bottleneck, and a decoder, with skip connections between corresponding encoder and decoder layers. Time embeddings are injected into each encoder block to condition the network on the current diffusion step.
	}
	\label{fig:unet-architecture}
\end{figure*}

\begin{figure*}[htbp]
	\centering
	\includegraphics[width=0.9\textwidth]{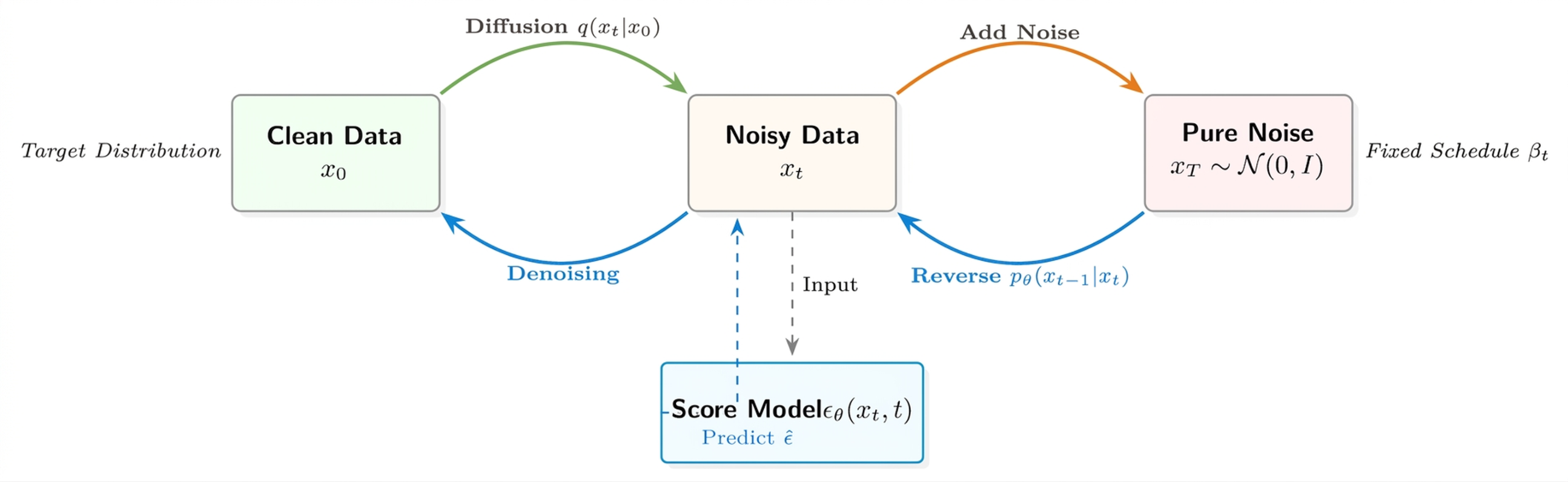}
	\caption{
		Illustration of the diffusion denoising process. In the forward process, Gaussian noise is gradually added to the clean data, transforming it into pure noise. In the reverse process, a 3D UNet is used iteratively to predict and remove the noise at each step, progressively reconstructing the clean data from the noisy input.
	}
	\label{fig:diffusion-denoising-process}
\end{figure*}

As shown in Fig.~\ref{fig:detailed_modules}, our proposed model leverages a 3D UNet-based diffusion architecture for denoising high-dimensional LIBS spectra. The 3D UNet is specifically designed to capture complex spatial, spectral, and sample-wise correlations, enabling effective noise suppression while preserving subtle and informative spectral features.

A key enhancement in our architecture is the integration of an attention mechanism within the feature extraction process, adaptively re-weighting learned features to focus on the most informative spectral regions while suppressing irrelevant components. This improves both denoising performance and feature quality for downstream regression tasks.
To improve generalization in few-shot scenarios, we incorporate feature group shuffling data augmentation, where encoded features are divided into groups and randomly permuted to generate augmented samples while preserving internal structure. This increases training data diversity and enables better capture of invariant patterns.
After denoising, attention-driven feature extraction, and data augmentation, the resulting latent representations can be directly used for quantitative regression (e.g., OLS or PLS) or further processed by an autoencoder for additional nonlinear dimensionality reduction, with modular design allowing flexible integration of different regression or classification heads.

\subsection{Mathematical Foundation of Diffusion Models}

Diffusion models are a class of generative models that learn to map between clean data and noise through a gradual, multi-step process. The forward process progressively adds Gaussian noise to the data over $T$ steps, transforming the original data distribution into a simple prior (typically standard normal). The reverse process aims to recover the clean data from noise by learning the conditional distributions at each step, parameterized by a neural network.
\subsubsection{Forward Diffusion Process}

The forward diffusion process gradually transforms the original data distribution $q(x_0)$ into a simple prior distribution (typically Gaussian noise) over $T$ steps. At each step $t$, we add a small amount of Gaussian noise according to a predefined schedule.

Given an original data sample $x_0 \sim q(x_0)$, the forward process produces a sequence $\{x_t\}_{t=1}^T$ by adding noise at each step:

\begin{equation}
	q(x_t | x_{t-1}) = \mathcal{N}(x_t; \sqrt{\alpha_t} x_{t-1}, (1 - \alpha_t) I)
\end{equation}

where $\alpha_t$ is a noise schedule parameter that controls the amount of noise added at step $t$. The cumulative effect of the forward process can be expressed as:

\begin{equation}
	q(x_t | x_0) = \mathcal{N}(x_t; \sqrt{\bar{\alpha}_t} x_0, (1 - \bar{\alpha}_t) I)
\end{equation}

where $\bar{\alpha}_t = \prod_{s=1}^t \alpha_s$. This allows us to sample $x_t$ directly from $x_0$ without computing the intermediate steps.

\subsubsection{Reverse Denoising Process}

The reverse process aims to learn the posterior distribution $q(x_{t-1} | x_t, x_0)$, which can be shown to be Gaussian:

\begin{equation}
	q(x_{t-1} | x_t, x_0) = \mathcal{N}(x_{t-1}; \tilde{\mu}_t(x_t, x_0), \tilde{\beta}_t I)
\end{equation}

where:
\begin{align}
	\tilde{\mu}_t(x_t, x_0) &= \frac{\sqrt{\bar{\alpha}_{t-1}} \beta_t}{1 - \bar{\alpha}_t} x_0 + \frac{\sqrt{\alpha_t} (1 - \bar{\alpha}_{t-1})}{1 - \bar{\alpha}_t} x_t \\
	\tilde{\beta}_t &= \frac{1 - \bar{\alpha}_{t-1}}{1 - \bar{\alpha}_t} \beta_t
\end{align}
\subsubsection{Noise Schedule}
The noise schedule determines how much Gaussian noise is added at each diffusion step. We use a linear schedule, where the noise variance $\beta_t$ increases linearly from $\beta_1$ to $\beta_T$ over $T$ steps. This gradual increase ensures a smooth transition from clean data to pure noise, facilitating stable training and effective denoising.
We use a linear noise schedule:
\begin{equation}
	\beta_t = \beta_1 + \frac{t - 1}{T - 1} (\beta_T - \beta_1)
\end{equation}

\subsubsection{Training Objective}

The diffusion model is trained to predict the noise $\epsilon$ that was added during the forward process. The training objective is to minimize the mean squared error between the predicted and true noise:

\begin{equation}
	\mathcal{L} = \mathbb{E}_{x_0, \epsilon, t} \left[ \| \epsilon_\theta(x_t, t) - \epsilon \|^2 \right]
\end{equation}

where $\epsilon_\theta$ is the neural network's prediction.

\subsection{Feature Extraction: 3D UNet}

The core of our diffusion model is a 3D UNet architecture, specifically designed for high-dimensional spectral data. It follows a U-shaped architecture with skip connections. The denoised output, $X_\text{denoised}$, serves as a high-quality input for subsequent analysis. This is particularly important for LIBS, where subtle spectral features may be distributed across multiple axes and traditional lower-dimensional models may fail to exploit such multi-way relationships. The 3D UNet structure thus provides a more expressive and generalizable feature extractor for high-dimensional, structured spectroscopic data.

\subsection{Attention}

In our model, an attention mechanism is integrated into the feature extraction process to enhance the network's ability to focus on informative spectral regions and suppress irrelevant or noisy components. Specifically, after each major encoding or decoding block, a channel-wise attention module is applied. This module consists of a small neural network that generates attention weights for each feature dimension, which are then used to adaptively re-weight the features. The attention mechanism enables the model to dynamically emphasize subtle but important spectral features, improving both denoising performance and the quality of the learned representations for downstream regression tasks.

Mathematically, given an input feature vector $\mathbf{h} \in \mathbb{R}^d$, the attention weights $\mathbf{a} \in \mathbb{R}^d$ are computed as:
\begin{equation}
	\mathbf{a} = \mathrm{Softmax}(W_2 \tanh(W_1 \mathbf{h} + \mathbf{b}_1) + \mathbf{b}_2)
\end{equation}
where $W_1, W_2$ are learnable weight matrices, $\mathbf{b}_1, \mathbf{b}_2$ are bias terms, and $\mathrm{Softmax}$ is applied element-wise. The output is the element-wise product of the attention weights and the input features:
\begin{equation}
	\mathbf{h}_{\text{att}} = \mathbf{a} \odot \mathbf{h}
\end{equation}
where $\odot$ denotes the Hadamard (element-wise) product. 

\subsection{Data Augmentation}

To further improve generalization and robustness, especially in few-shot scenarios, we employ a data augmentation strategy based on feature group shuffling. After that, each sample's feature vector is divided into several groups of equal size. The order of these groups is randomly permuted to generate new, augmented samples, while the internal structure of each group is preserved. This process is repeated multiple times to expand the training set. The corresponding target value for each augmented sample remains unchanged. This augmentation increases the diversity of the training data, mitigates overfitting, and helps the model learn invariant and discriminative patterns across different samples.

\subsection{Regression}

In our experiments, we employ ordinary least squares (OLS) regression, although other regression models such as partial least squares (PLS) or support vector regression (SVR) can also be used. The regression model is trained to map the extracted features to the target values (e.g., elemental concentrations). Model performance is evaluated using standard metrics such as mean squared error (MSE), root mean squared error (RMSE), mean absolute error (MAE), and $R^2$ score, providing a comprehensive assessment of prediction accuracy and generalization ability.

\section{Implementation and Results}
\subsection{Dataset and Environment}

\begin{figure}[htbp]
	\centering
	\includegraphics[width=0.8\linewidth]{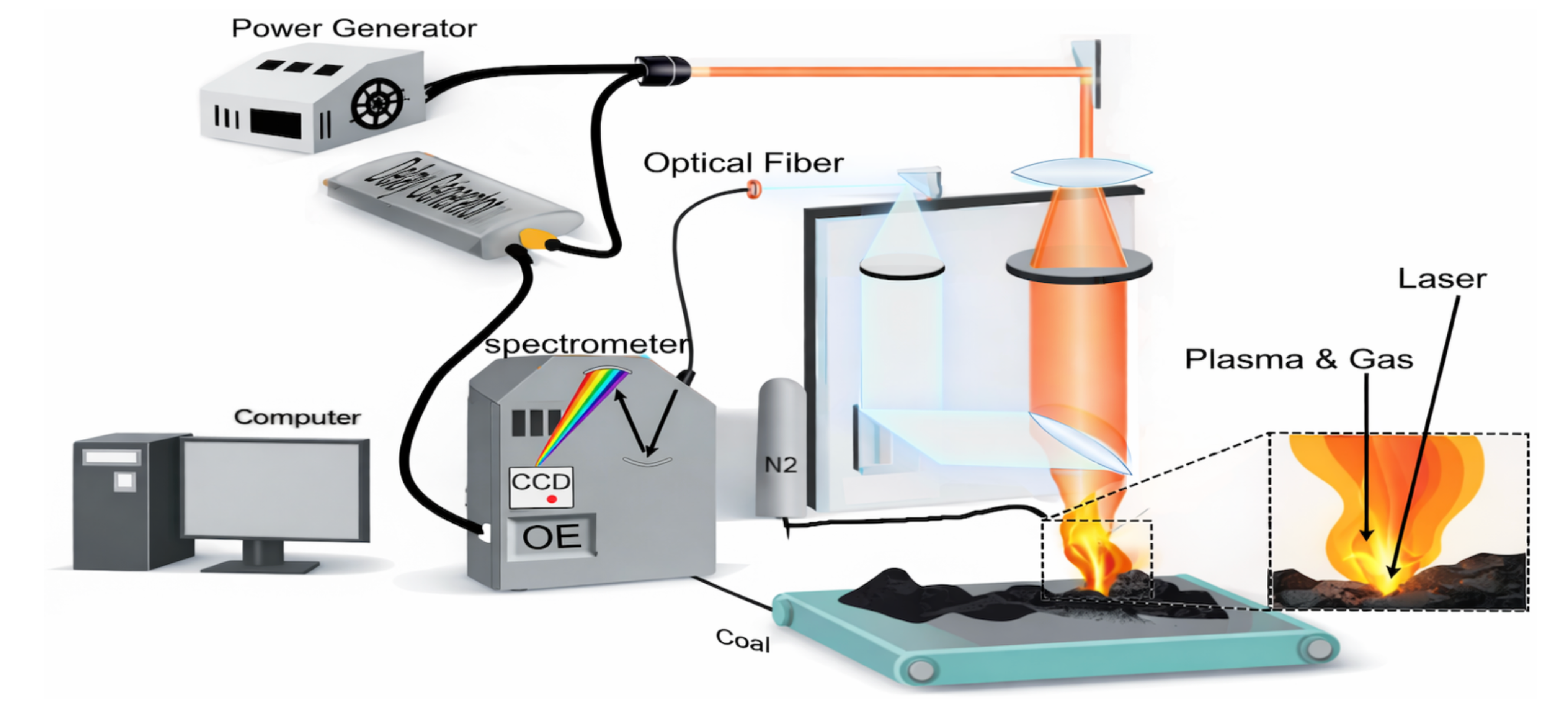}
	\caption{Schematic diagram of the LIBS system}
	\label{fig:libs_system}
\end{figure}

The dataset was collected using a custom-built LIBS system designed for coal quality analysis on a transport belt, as illustrated in Figure~\ref{fig:libs_system}. The system employs a 1064 nm pulsed Nd:YAG laser (100 mJ pulse energy, 8 ns pulse width, 10 Hz repetition rate) directed through a dichroic mirror and focused onto the coal surface using a plano-convex lens (focal length: 100 mm), creating a laser spot diameter of approximately 500 $\mu$m. The resulting plasma emission is collected by the same lens, reflected by the dichroic mirror, and transmitted via a 600 $\mu m$ core diameter optical fiber to an Echelle spectrometer (Andor Mechelle ME5000) with a spectral range of 180-800 nm and resolution of 0.1 nm.

All programming experiments were conducted on a workstation equipped with NVIDIA RTX 4090 GPU (16 GB VRAM), Intel Core i9-13980HX Processor (64 GB DDR5), and software environments including CUDA Version 12.1, Python Version 3.10, and PyTorch Version 2.0.1.

\subsection{Data Preprocessing}
Due to the high dimensionality of LIBS spectral data and GPU memory constraints, we implemented several strategies to address memory bottlenecks during diffusion denoising: (1) \textbf{Block Processing}: Data is divided into smaller blocks (4096 samples per block) and processed sequentially; (2) \textbf{Automatic Padding}: Input dimensions are padded to ensure compatibility with downsampling/upsampling operations; (3) \textbf{Gradient Accumulation}: Gradients are accumulated over multiple mini-batches before optimizer steps, simulating larger batch sizes without increasing memory usage.

\subsection{Ablation Study}

We conducted an ablation study comparing four approaches: Diffusion-DA-AE (complete pipeline), DA-AE (without diffusion denoising), AE (without data augmentation), and PCA-PLS (traditional baseline). As shown in Table~\ref{tab:ablation_study}, the diffusion denoising module shows significant improvements, particularly for low-concentration elements where noise impact is critical. While traditional methods can be competitive for simple spectral signatures, our deep learning approach provides superior performance for complex, non-linear relationships. The complete Diffusion-DA-AE pipeline achieves the best overall performance, demonstrating that each component contributes meaningfully to the final results.

\begin{table}[htbp]
	\centering
	\caption{Ablation study results: Mean performance metrics across 21 elemental concentrations}
	\begin{tabular}{lccccc}
		\toprule
		Method & MSE & RMSE & RMAE & MAE & MRE \\
		\midrule
		Diffusion & 18.25 & 2.85 & 0.2847 & 9.85 & 0.95 \\
		DA-AE & 19.05 & 3.12 & 0.3012 & 9.95 & 1.02 \\
		AE & 51.45 & 5.67 & 0.4123 & 10.15 & 1.35 \\
		PCA-PLS & 85.65 & 6.85 & 0.4568 & 13.45 & 2.85 \\
		\bottomrule
	\end{tabular}
	\label{tab:ablation_study}
\end{table}

\subsection{Denoising Performance}

To comprehensively evaluate the denoising performance of different methods, we compare the number of  mean peak height, mean signal-to-noise ratio (SNR) for each method.

\begin{table}[htbp]
	\centering
	\includegraphics[width=0.9\linewidth]{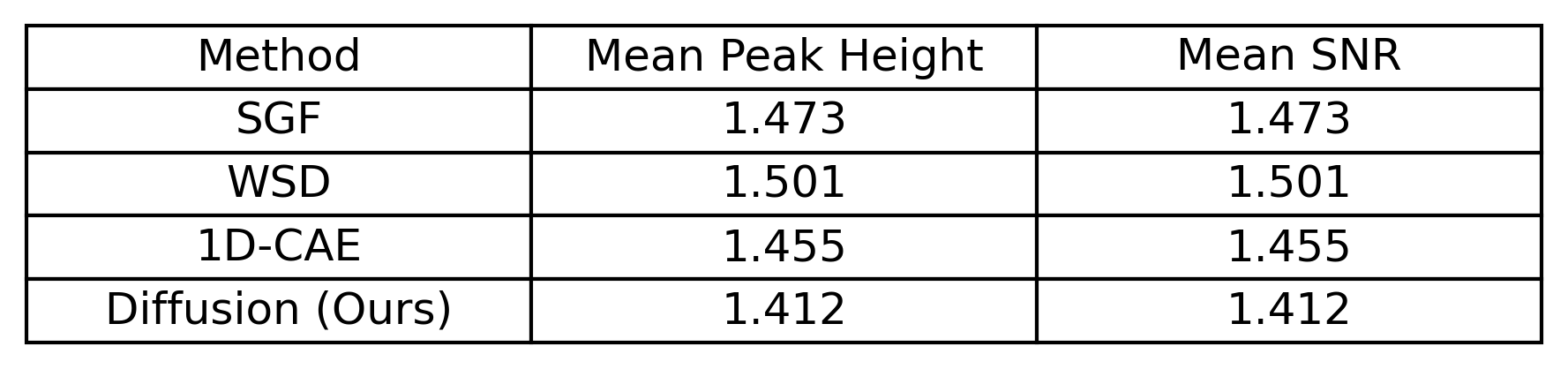}
	\caption{Mean peak height and mean SNR of detected peaks.}
	\label{tab:snr}
\end{table}

Table~\ref{tab:snr} demonstrates that the mean peak height and mean SNR of the diffusion-based method are comparable to those of traditional methods, indicating that the preserved peaks are not only numerous but also have high signal quality. Importantly, the number of unique peaks detected by the diffusion model is significantly higher than that of traditional methods, suggesting superior detail preservation and the ability to recover weak but meaningful spectral features.

\subsection{Attention VS Self-regression}

\begin{figure}[htbp]
	\centering
	\begin{minipage}[t]{0.48\linewidth}
		\centering
		\includegraphics[width=\linewidth]{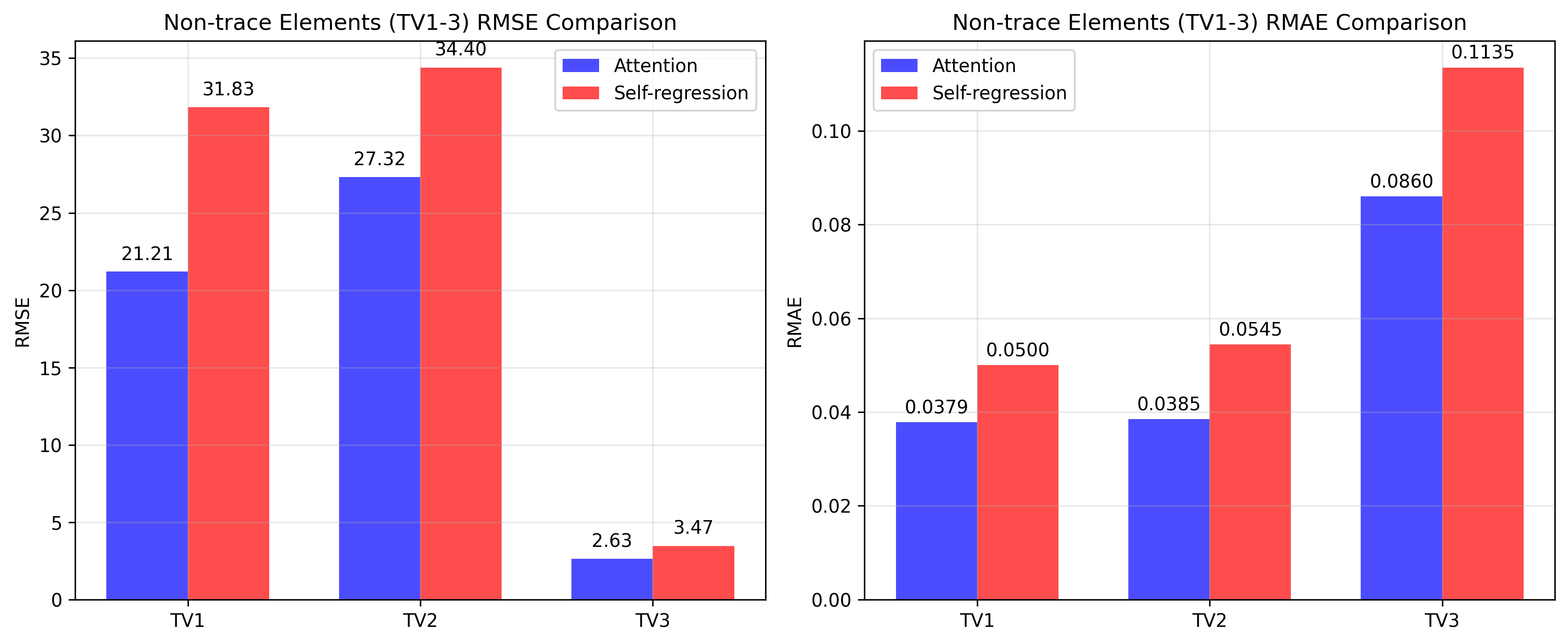}
	\end{minipage}
	\hfill
	\begin{minipage}[t]{0.48\linewidth}
		\centering
		\includegraphics[width=\linewidth]{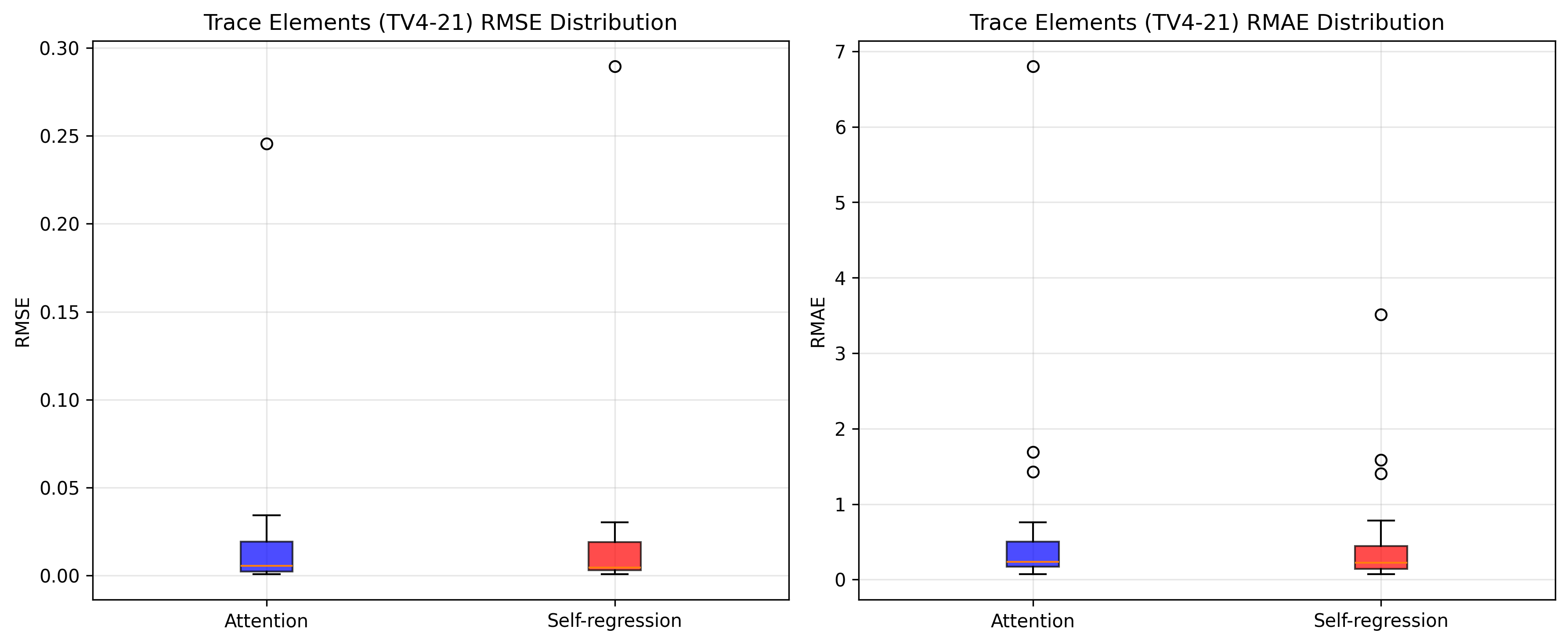}
	\end{minipage}
	\caption{Performance comparison between Attention and Self-regression mechanisms for major elements (left) and trace elements (right). Attention shows 20.6-33.4\% RMSE and 24.2-32.2\% RMAE improvements for major elements.}
	\label{fig:performance_comparison}
\end{figure}

We compared attention-based and self-regression-based feature extraction mechanisms in our autoencoder architecture. Attention mechanisms adaptively re-weight learned features to focus on informative spectral regions, while self-regression mechanisms learn to predict features from their own representations. Experimental results show distinct performance patterns between major elements (TV1-3) and trace elements (TV4-21), as illustrated in Figure~\ref{fig:performance_comparison}, with attention mechanisms demonstrating superior performance for major elements, achieving 33.4\% improvement in RMSE for TV1 (21.21 vs 31.83) compared to self-regression approaches.

\subsection{Data Augmentation}

\begin{figure}[htbp]
	\centering
	\includegraphics[width=0.8\linewidth]{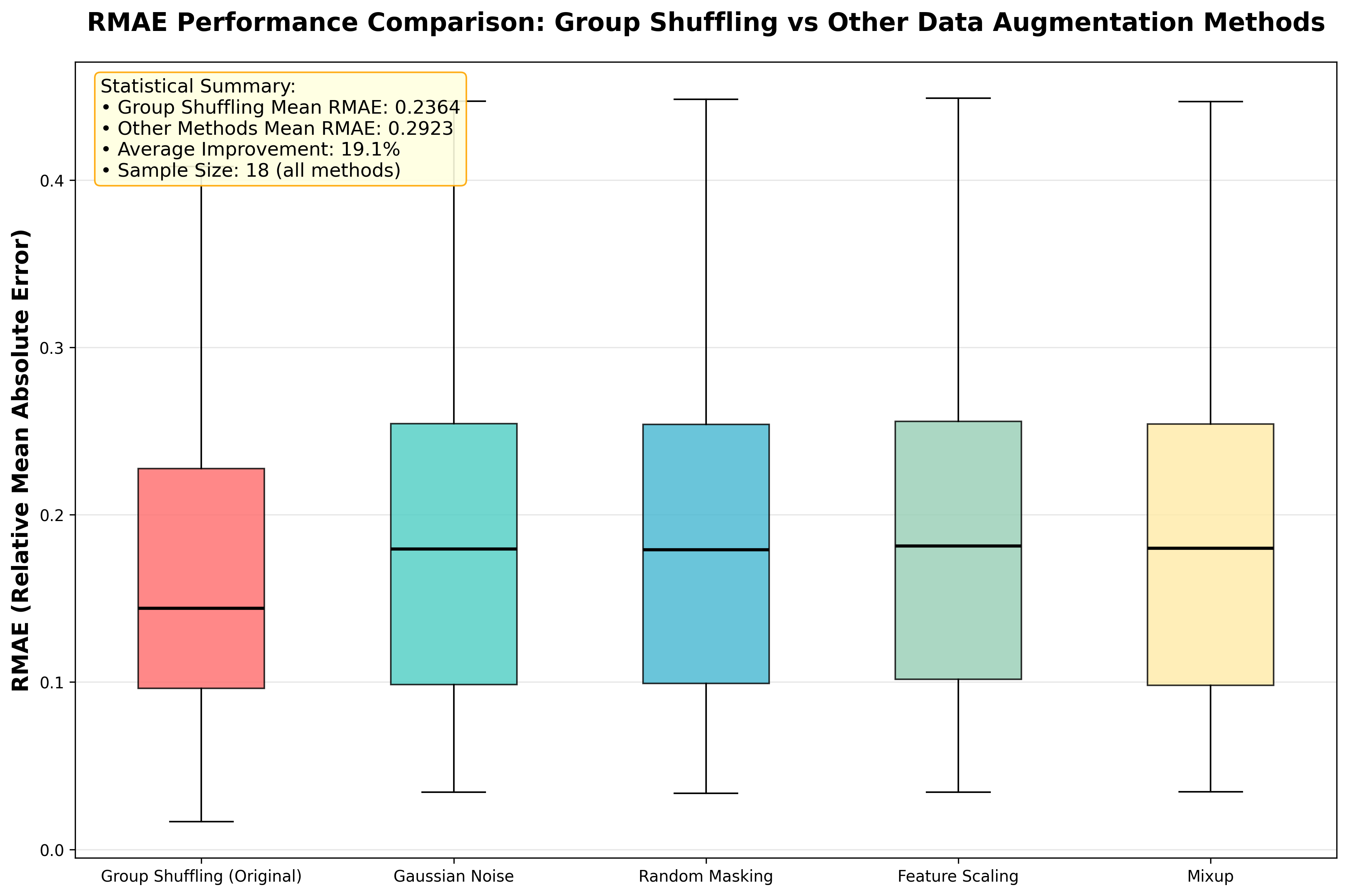}
	\caption{RMAE performance comparison between Group Shuffling and other data augmentation methods.}
	\label{fig:rmae_boxplot_colors}
\end{figure}

We systematically evaluated the effectiveness of Group Shuffling augmentation by comparing it with four widely used methods (Gaussian Noise, Random Masking, Feature Scaling, and Mixup) on 21 true-value samples.
As shown in Figure~\ref{fig:rmae_boxplot_colors}, Group Shuffling achieved the lowest mean RMAE (0.2364) compared to other methods (0.2923), representing a 19.1\% improvement, with narrower interquartile range (0.10-0.23) indicating more stable and consistent performance across different true values.

\section{Conclusion}
This work presents a standardized processing pipeline for high-accuracy few-shot regression in laser-induced breakdown spectroscopy (LIBS). Our pipeline demonstrates significant improvements over traditional methods, addressing the critical challenges of spectral noise and data scarcity in resource-constrained environments.The proposed pipeline represents a significant step forward in addressing the fundamental challenges of LIBS analysis in few-shot scenarios. 

{\small
	\bibliographystyle{IEEEtran}
	\bibliography{refs}

@article{yan2025review,
	title={A review on spectral data preprocessing techniques for machine learning and quantitative analysis},
	author={Yan, Chunsheng},
	journal={iScience},
	year={2025},
	publisher={Elsevier}
}

@article{eilers2003perfect,
	title={A perfect smoother},
	author={Eilers, Paul HC},
	journal={Analytical chemistry},
	volume={75},
	number={14},
	pages={3631--3636},
	year={2003},
	publisher={ACS Publications}
}

@article{savitzky1964smoothing,
	title={Smoothing and differentiation of data by simplified least squares procedures.},
	author={Savitzky, Abraham and Golay, Marcel JE},
	journal={Analytical chemistry},
	volume={36},
	number={8},
	pages={1627--1639},
	year={1964},
	publisher={ACS Publications}
}

@article{shen2022single,
	title={Single convolutional neural network model for multiple preprocessing of Raman spectra},
	author={Shen, Jiahao and Li, Miao and Li, Zhongfeng and Zhang, Zhuoyong and Zhang, Xin},
	journal={Vibrational Spectroscopy},
	volume={121},
	pages={103391},
	year={2022},
	publisher={Elsevier}
}

@article{kazemzadeh2022cascaded,
	title={Cascaded deep convolutional neural networks as improved methods of preprocessing raman spectroscopy data},
	author={Kazemzadeh, Mohammadrahim and Martinez-Calderon, Miguel and Xu, Weiliang and Chamley, Lawrence W and Hisey, Colin L and Broderick, Neil GR},
	journal={Analytical Chemistry},
	volume={94},
	number={37},
	pages={12907--12918},
	year={2022},
	publisher={ACS Publications}
}

@article{jiao2024three,
	title={A three-stage deep learning-based training frame for spectra baseline correction},
	author={Jiao, Qingliang and Cai, Boyong and Liu, Ming and Dong, Liquan and Hei, Mei and Kong, Lingqin and Zhao, Yuejin},
	journal={Analytical Methods},
	volume={16},
	number={10},
	pages={1496--1507},
	year={2024},
	publisher={Royal Society of Chemistry}
}

@article{he2023new,
	title={A new technique for baseline calibration of soil X-ray fluorescence spectra based on enhanced generative adversarial networks combined with transfer learning},
	author={He, Xinghua and Zhao, Yanchun and Li, Fusheng},
	journal={Journal of Analytical Atomic Spectrometry},
	volume={38},
	number={11},
	pages={2486--2498},
	year={2023},
	publisher={Royal Society of Chemistry}
}

@article{shang2024study,
	title={Study on breast cancerization and isolated diagnosis in situ by HOF-ATR-MIR spectroscopy with deep learning},
	author={Shang, Hui and Wu, Qingxia and Wu, Jinjin and Zhou, Suwei and Wang, Zihan and Wang, Huijie and Yin, Jianhua},
	journal={Spectrochimica Acta Part A: Molecular and Biomolecular Spectroscopy},
	volume={319},
	pages={124546},
	year={2024},
	publisher={Elsevier}
}

@article{vrabel2020restricted,
	title={Restricted Boltzmann Machine method for dimensionality reduction of large spectroscopic data},
	author={Vr{\'a}bel, J and Po{\v{r}}{\'\i}zka, P and Kaiser, J},
	journal={Spectrochimica Acta Part B: Atomic Spectroscopy},
	volume={167},
	pages={105849},
	year={2020},
	publisher={Elsevier}
}

@article{yuan2021rapid,
	title={Rapid classification of steel via a modified support vector machine algorithm based on portable fiber-optic laser-induced breakdown spectroscopy},
	author={Yuan, Mengtian and Zeng, Qingdong and Wang, Jie and Li, Wenxin and Chen, Guanghui and Li, Zitao and Liu, Yang and Guo, Lianbo and Li, Xiangyou and Yu, Huaqing},
	journal={Optical Engineering},
	volume={60},
	number={12},
	pages={124114--124114},
	year={2021},
	publisher={Society of Photo-Optical Instrumentation Engineers}
}

@article{ma2023step,
	title={A step-by-step classification method of coal and miscellaneous materials by laser-induced breakdown spectroscopy},
	author={Ma, Weizhe and Yu, Ziyu and Lu, Zhimin and Ma, Qingxiang and Yao, Shunchun},
	journal={At. Spectrosc},
	volume={44},
	number={3},
	pages={160--168},
	year={2023}
}

@article{li2023high,
	title={High-accuracy measurement of the heat of detonation with good robustness by laser-induced breakdown spectroscopy of energetic materials},
	author={Li, An and Zhang, Xinyu and Yin, Yunsong and Wang, Xianshuang and He, Yage and Shan, Yuheng and Zhang, Ying and Liu, Xiaodong and Zhong, Lixiang and Liu, Ruibin},
	journal={Journal of Analytical Atomic Spectrometry},
	volume={38},
	number={4},
	pages={810--817},
	year={2023},
	publisher={Royal Society of Chemistry}
}

@article{li2023real,
	title={Real time and high-precision online determination of main components in iron ore using spectral refinement algorithm based LIBS},
	author={Li, An and Zhang, Xinyu and Liu, Xiaodong and He, Yage and Shan, Yuheng and Sun, Haohan and Yi, Wen and Liu, Ruibin},
	journal={Optics Express},
	volume={31},
	number={23},
	pages={38728--38743},
	year={2023},
	publisher={Optica Publishing Group}
}

@article{zhang2024determination,
	title={Determination of propellant products by time resolved and spatial distribution LIPS combined with high-speed schlieren imaging},
	author={Zhang, Xinyu and Li, An and Zhang, Ying and Yin, Yunsong and Wang, Xianshuang and He, Yage and Lyv, Jing and Shan, Yuheng and Liu, Xiaodong and Yi, Wen and others},
	journal={Journal of Analytical Atomic Spectrometry},
	volume={39},
	number={3},
	pages={974--981},
	year={2024},
	publisher={Royal Society of Chemistry}
}

@article{van2023deep,
	title={Deep learning regression for quantitative LIBS analysis},
	author={Van den Eynde, Simon and Diaz-Romero, Dillam Jossue and Zaplana, Isiah and Peeters, Jef},
	journal={Spectrochimica Acta Part B: Atomic Spectroscopy},
	volume={202},
	pages={106634},
	year={2023},
	publisher={Elsevier}
}

@article{ctvrtnickova2009,
	title={Characterization of coal fly ash components by laser-induced breakdown spectroscopy},
	author={Ctvrtnickova, Tereza and Mateo, Mari-Paz and Yanez, Armando and Nicolas, Gines},
	journal={Spectrochimica Acta Part B: Atomic Spectroscopy},
	volume={64},
	number={10},
	pages={1093--1097},
	year={2009},
	publisher={Elsevier}
}

@article{boucher2015study,
	title={A study of machine learning regression methods for major elemental analysis of rocks using laser-induced breakdown spectroscopy},
	author={Boucher, Thomas F and Ozanne, Marie V and Carmosino, Marco L and Dyar, M Darby and Mahadevan, Sridhar and Breves, Elly A and Lepore, Kate H and Clegg, Samuel M},
	journal={Spectrochimica Acta Part B: Atomic Spectroscopy},
	volume={107},
	pages={1--10},
	year={2015},
	publisher={Elsevier}
}

@article{shahani2022machine,
	title={Machine learning-based intelligent prediction of elastic modulus of rocks at thar coalfield},
	author={Shahani, Niaz Muhammad and Zheng, Xigui and Guo, Xiaowei and Wei, Xin},
	journal={Sustainability},
	volume={14},
	number={6},
	pages={3689},
	year={2022},
	publisher={MDPI}
}

@article{he2023tdiffde,
	title={Tdiffde: A truncated diffusion model for remote sensing hyperspectral image denoising},
	author={He, Jiang and Li, Yajie and Yuan, Qiangqiang and others},
	journal={arXiv preprint arXiv:2311.13622},
	year={2023}
}

@inproceedings{miao2023dds2m,
	title={DDS2M: Self-supervised denoising diffusion spatio-spectral model for hyperspectral image restoration},
	author={Miao, Yuchun and Zhang, Lefei and Zhang, Liangpei and Tao, Dacheng},
	booktitle={Proceedings of the IEEE/CVF International Conference on Computer Vision},
	pages={12086--12096},
	year={2023}
}

@article{jiang2021baseline,
	title={Baseline correction method based on improved adaptive iteratively reweighted penalized least squares for the x-ray fluorescence spectrum},
	author={Jiang, Xiaoyu and Li, Fusheng and Wang, Qingya and Luo, Jie and Hao, Jun and Xu, Muqiang},
	journal={Applied Optics},
	volume={60},
	number={19},
	pages={5707--5715},
	year={2021},
	publisher={Optical Society of America}
}

@article{donoho1994ideal,
	title={Ideal spatial adaptation by wavelet shrinkage},
	author={Donoho, David L and Johnstone, Iain M},
	journal={biometrika},
	volume={81},
	number={3},
	pages={425--455},
	year={1994},
	publisher={Oxford University Press}
}

@article{kalman1960new,
	title={A new approach to linear filtering and prediction problems},
	author={Kalman, Rudolph Emil},
	year={1960}
}

@article{gilda2019automatic,
	title={Automatic Kalman-filter-based wavelet shrinkage denoising of 1D stellar spectra},
	author={Gilda, Sankalp and Slepian, Zachary},
	journal={Monthly Notices of the Royal Astronomical Society},
	volume={490},
	number={4},
	pages={5249--5269},
	year={2019},
	publisher={Oxford University Press}
}

@article{qiao2018protein,
	title={Protein-protein interface hot spots prediction based on a hybrid feature selection strategy},
	author={Qiao, Yanhua and Xiong, Yi and Gao, Hongyun and Zhu, Xiaolei and Chen, Peng},
	journal={BMC bioinformatics},
	volume={19},
	number={1},
	pages={14},
	year={2018},
	publisher={Springer}
}

@article{liu2025learning,
	title={Learning to Decide with Just Enough: Information-Theoretic Context Summarization for CMDPs},
	author={Liu, Peidong and Lin, Junjiang and Wang, Shaowen and Xu, Yao and Li, Haiqing and Xie, Xuhao and Wu, Siyi and Li, Hao},
	journal={arXiv preprint arXiv:2510.01620},
	year={2025}
}

@inproceedings{li2021latency,
	title={Latency-aware batch task offloading for vehicular cloud: Maximizing submodular bandit},
	author={Li, Hao and Huang, Haitao and Qian, Zhuzhong},
	booktitle={2021 IEEE 14th International Conference on Cloud Computing (CLOUD)},
	pages={584--593},
	year={2021},
	organization={IEEE}
}

@inproceedings{xu2017reliable,
	title={A reliable resource scheduling for network function virtualization},
	author={Xu, Daoqiang and Li, Yefei and Yin, Ming and Li, Xin and Li, Hao and Qian, Zhuzhong},
	booktitle={International Conference on Security, Privacy and Anonymity in Computation, Communication and Storage},
	pages={251--260},
	year={2017},
	organization={Springer}
}

@article{li2026golden,
	title={Golden RPG: Confidence-Adaptive Region-Aware Noise for Compositional Text-to-Image Generation},
	author={Li, Hao},
	journal={arXiv preprint arXiv:2604.25314},
	year={2026}
}

@article{li2026revisiting,
	title={Revisiting the Scale Loss Function and Gaussian-Shape Convolution for Infrared Small Target Detection},
	author={Li, Hao and Zhuo, Man Fung},
	journal={arXiv preprint arXiv:2604.09991},
	year={2026}
}

@article{li2026r,
	title={R3D: Regional-guided Residual Radar Diffusion},
	author={Li, Hao and Liu, Xinqi and Jin, Yaoqing},
	journal={arXiv preprint arXiv:2601.06465},
	year={2026}
}

@article{li2026multi,
	title={Multi-Adapter PPO: A Cross-Attention Enhanced Wavelength Selection Framework for LIBS Quantitative Analysis},
	author={Li, Hao and Zhuo, Man Fung},
	journal={arXiv preprint arXiv:2606.17476},
	year={2026}
}

@article{harefa2021performing,
	title={Performing sequential forward selection and variational autoencoder techniques in soil classification based on laser-induced breakdown spectroscopy},
	author={Harefa, Edward and Zhou, Weidong},
	journal={Analytical Methods},
	volume={13},
	number={41},
	pages={4926--4933},
	year={2021},
	publisher={The Royal Society of Chemistry}
}

@article{ding2018hybrid,
	title={A Hybrid Feature Selection Algorithm Based on Information Gain and Sequential Forward Floating Search},
	author={Ding, Jianli and Fu, Liyang},
	journal={J Intell Comput},
	volume={9},
	number={3},
	pages={93},
	year={2018}
}

@article{wei2020variations,
	title={Variations in variational autoencoders-a comparative evaluation},
	author={Wei, Ruoqi and Garcia, Cesar and El-Sayed, Ahmed and Peterson, Viyaleta and Mahmood, Ausif},
	journal={Ieee Access},
	volume={8},
	pages={153651--153670},
	year={2020},
	publisher={IEEE}
}
}

\end{document}